\documentclass{article}
\usepackage[preprint]{neurips_2023}
\usepackage[utf8]{inputenc} 
\usepackage[T1]{fontenc}    
\usepackage{hyperref}       
\usepackage{graphicx}
\usepackage{url}            
\usepackage{booktabs}       
\usepackage{amsfonts}       
\usepackage{nicefrac}       
\usepackage{microtype}      
\usepackage{natbib}

\title{Knowledge-Centric Metacognitive Learning}

\newif\ifuniqueAffiliation
\uniqueAffiliationtrue

\ifuniqueAffiliation 
\author{Arun Kumar \\
	Dept. of Computer Science \\
	University of Minnesota, Twin Cities \\
	Minneapolis, MN, USA \\
	\texttt{kumar250@umn.edu} \\
	\And
	Paul Schrater \\
	Depts. of Computer Science and Psychology \\
	University of Minnesota, Twin Cities \\
	Minneapolis, MN, USA \\
	\texttt{schrater@umn.edu} \\
}
\fi

\hypersetup{
pdftitle={kix},
pdfauthor={Arun Kumar, Paul Schrater},
}
\graphicspath{{Figures/}}
\usepackage{balance}
\usepackage{subcaption}
\usepackage{wrapfig}

\usepackage{amsmath}
\usepackage{amsfonts}
\usepackage{amssymb}
\usepackage{mathtools}
\usepackage{booktabs}
\usepackage{stmaryrd}

\usepackage[ruled,lined]{algorithm2e}

\usepackage{enumitem}
\usepackage{cleveref}       

\DontPrintSemicolon

\usepackage{xcolor}

\usepackage{dsfont}
\usepackage{bbm, bbold}
\usepackage{tabularx}

\newcommand{\Comment}{\tcc*[r]}
\SetKwComment{tcc}{$\triangleright$~}{}
\SetKwComment{Comment}{/* }{ */}
\SetCommentSty{normalfont}
\SetCommentSty{mycommfont}
\SetKwInput{KwRequire}{Input}
\SetNlSty{}{}{:}

\begin{document}
\maketitle
\begin{abstract}
Interactions are central to intelligent reasoning and learning abilities, with the interpretation of abstract knowledge guiding meaningful interaction with objects in the environment. While humans readily adapt to novel situations by leveraging abstract knowledge acquired over time, artificial intelligence systems lack principled mechanisms for incorporating abstract knowledge into learning, leading to fundamental challenges in the emergence of intelligent and adaptive behavior. To address this gap, we introduce knowledge-centric metacognitive learning based on three key principles: natural abstractions, knowledge-guided interactions through interpretation, and the composition of interactions for problem solving. Knowledge learning facilitates the acquisition of abstract knowledge and the association of interactions with knowledge, while object interactions guided by abstract knowledge enable the learning of transferable interaction concepts, abstract reasoning, and generalization. This metacognitive mechanism provides a principled approach for integrating knowledge into reinforcement learning and offers a promising pathway toward intelligent and adaptive behavior in artificial intelligence, robotics, and autonomous systems.
\end{abstract}
\section*{Introduction}
\label{sec:introduction}
A hallmark of human intelligence is the capacity to construct and interpret abstract knowledge. Interpretation grounded in such abstractions enables meaningful interaction with the environment, forming the foundation for knowledge learning, flexible planning, and decision-making. It is fascinating how effortlessly people discern natural abstractions and leverage them to interact with their environments in flexible and adaptive ways. In contrast, contemporary artificial intelligence systems lack the principled integration of abstract knowledge into learning, resulting in fundamental limitations in the emergence of intelligent behavior.
For example, autonomous systems trained solely on object-level data often fail to respond to changes in world characteristics due to the absence of integrated knowledge learning. Such gaps in knowledge learning lead to non-robust, unreliable, and specialized behavior in unknown and complex scenarios.
Machine learning methods \cite{Goodfellow2016,bishop2006pattern} learn a mapping function by optimizing an objective between predictions and targets; however, these targets lack the underlying know-how required for world understanding. In reinforcement learning \cite{sutton2018reinforcement}, agents collect experiences through interaction with their environments and learn policies for action selection, yet even model-based approaches, despite learning a dynamics model, fail to integrate abstract knowledge into learning and world models. Learning, at its core, is \emph{knowledge learning} for an abstract understanding of the world and \emph{responding adaptively} through the use of such knowledge. 
Addressing these gaps necessitates a knowledge-centric metacognitive learning approach.

\begin{figure}[!ht]
        \centering
        \begin{subfigure}[b]{0.445\linewidth}
            \centering
            \includegraphics[width=0.99\textwidth]{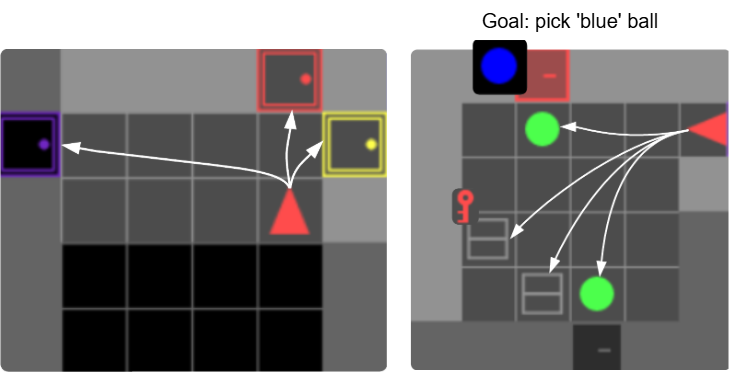}
            \caption{Overview}
        \end{subfigure}
        \begin{subfigure}[b]{0.545\linewidth}
            \centering
            \includegraphics[width=0.99\textwidth]{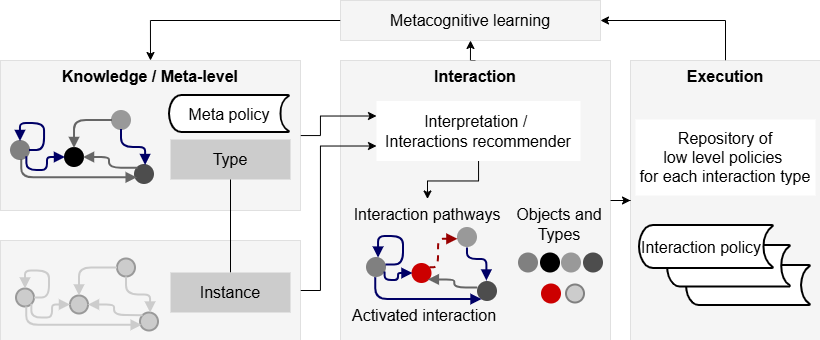}
            \caption{Approach}
        \end{subfigure}
    \caption{
    a) Illustration of metacognitive approach in a navigation domain, where an agent must develop an abstract understanding of its environment and interpret it to achieve the goal of retrieving a blue ball located behind a locked door, blocked by obstacles, with the key hidden inside a box. 
    b) A knowledge centric metacognitive learning approach exploits inherent object abstractions and relations to generate interaction goals and compose interactions, thereby enabling the emergence of intelligent and flexible behavior.
    }
    \label{fig:kix_overview}
    \vspace*{-3mm}
\end{figure}

Meta-reasoning about subsequent actions underpins flexible decision-making. More broadly, meta-reasoning acts as a sense-making process within knowledge systems, enabling people to intuitively direct their learning to discern abstract relational structure and ground object interactions. However, meta-reasoning in artificial systems predominantly acts as a resource controller, monitoring lower-level resources and managing resource allocation for efficient functioning. While such resource management is essential, a resource-controller–only view is severely limiting for facilitating abstract reasoning and adaptive behaviors. Meta-reasoning must pivot toward interpretation-based reasoning within abstract knowledge. This paradigm of abstract interpretive reasoning and learning -- termed \emph{metacognitive reasoning and learning} -- is central to much of knowledge learning, interaction with objects in the physical world, and manifested in the emergence of autonomous intelligent behaviors. Figure~\ref{fig:kix_overview} illustrates this process in a navigation domain. To develop an understanding of the world, an agent must learn abstract knowledge of objects where object types serve as natural abstractions, their relations, and the ways in which they can be interacted with. Interpretation over this abstract knowledge guides interactions enabling flexible problem solving.

%

Our proposal is based on three main principles for knowledge-centric learning and problem-solving.
\emph{Natural abstraction}: 
Real-world environments are naturally composed of entities categorized by their types, similar to an inherent hierarchy of objects and classes. Objects in the {\em instance space} inherit properties and relations from a higher-level {\em type space}. Type space graphs provide an abstract representation of objects and their relationships within the environment. The abstraction space is where the knowledge resides and is referred to as the knowledge or meta level.
\emph{Interactions}: 
An agent’s metacognitive process employs type space graphs to create plans that specify which entities to interact with and the nature of these interactions in order to achieve its goals. While the meta-level policy guides the agent based on abstract knowledge, the agent executes interactions in the {\em instance space}, where objects are type instantiations. This constitutes the interaction level.
\emph{Composition of interactions}: 
The meta-level guides the agent to interact with recommended objects using interaction types, and the agent executes low-level policies corresponding to guided interactions. The composition of these interactions forms interaction pathways for graph traversal at both the meta and instance level, thereby facilitating the execution of low-level actions. The process is loosely coupled through interactions, enabling flexible and adaptive behavior.

Integrating these principles into metacognitive learning provides foundational mechanisms for endowing agents with knowledge learning and adaptive capabilities, arising from their ability to interpret abstract knowledge, learn object interactions guided by meta-level policies, and leverage interaction pathways for adaptation. The proposed metacognitive learning approach captures the essence of abstract reasoning, knowledge learning, and flexible behavior, with potentially far-reaching implications across domains that require abstract reasoning and interaction with the physical world.
In the remaining of this article, we describe the key components and exemplifying characteristics of the approach, followed by an experimental evaluation in a navigation domain, and discuss how it enables adaptive problem solving through knowledge–guided interaction learning.
\section*{Background}
\label{sec:kix_background}
In this section, we briefly describe closely related work and the general setting of meta-reasoning, graph neural networks, and policy gradient methods.
\subsection*{Related Work}
Compositionality and abstractions are central to generalization, with abstraction requiring a mechanism for grounding representations in objects. 
An object-centric model \cite{veerapaneni2020entity} was presented in which object representations were constructed from raw observations and local functions with uniformity were then used for prediction and planning. A symbolic object-centric representation \cite{vijay2019generalization} was used to study the impact of adding prior knowledge through relations in an object graph. 
These studies focus primarily on object space, which inhibits the abstraction of objects and relational structures naturally occurring in the real world and used for decision-making. 
Key questions include what a natural abstraction should be and how it can be used to compose interactions as building blocks for problem solving. We address these largely overlooked questions by preserving abstract knowledge and relational higher-order structures in the form of knowledge graphs, where abstractions are grounded using type space. Learning a compositional structure is achieved through learning over knowledge graphs with neighborhood aggregation. 

Hierarchical reinforcement learning approaches \cite{dayan1992feudal,sutton1999between,pateria2021hierarchical,bagaria2021skill} learn a higher-level policy by choosing subtasks as higher-level actions, which decompose tasks into subtasks. Higher-level actions persist over a longer timescale, referred to as temporal abstraction \cite{barto2003recent,dietterich2000hierarchical,sutton1999between}. In contrast to options \cite{sutton1999between}, which combine a subtask with the core MDP, MAXQ value function decomposition \cite{dietterich2000hierarchical} decomposes the core MDP into smaller sub-MDPs, each with a separate learnable policy. Other methods induce learning hierarchies based on teacher signals \cite{portelas2020teacher,hutsebaut2022hierarchical} or a progression from easier to more difficult tasks \cite{campero2020learning}. These methods enable agents to learn hierarchy of policies, but they remain in the object space for both low- and high-level actions and do not meet key meta-reasoning demands: type abstractions and loose coupling. Methods such as \cite{ten2022curiosity,oudeyer2007intrinsic} encourage exploration through intrinsic reward or motivation, \cite{klyubin2005empowerment} develop internal goals for empowerment, focusing on aspects different than ours. 

Goal-conditioned reinforcement learning \cite{liu2022goal,pong2018temporal} seeks to support multi-task generalization by conditioning policies on goals, yet remains challenged due to the lack of integrated knowledge based reasoning and learning in complex tasks. We address these limitations through a metacognitive approach, in which abstract knowledge constrains learning, generates and guides interactions, enabling flexible behavior.
Neuro-symbolic approaches \cite{hitzler2022neuro,zhang2024neuro} aim to overcome the limitations of purely symbolic or neural methods by integrating them into hybrid models that combine the learning capacity of neural networks with the logical inference and interpretability of symbolic systems. However, significant challenges persist in facilitating interaction and integration between these disparate paradigms to ensure a seamless and semantically grounded unification. 

Our approach addresses these demands by leveraging natural higher-order knowledge structures, represented as type-space graphs, for meta-level policy learning, and by mediating object interactions via an intermediary between the meta and execution levels, thus enabling loose coupling and reuse of low-level policies and interaction policies can be swapped in and out based on interaction type. Our work is distinguished by its principled incorporation of relational abstract knowledge to guide interactions that govern low-level execution.
\subsection*{Meta-reasoning}
Meta-reasoning is often defined as the set of processes that monitor our thought processes and regulate resource allocation \cite{ackermanthompson2017meta}. Its basic structure \cite{nelson1990metamemory} consists of levels: meta and object. The meta level's functions include control and monitoring. The object's state changes as control passes from the meta to the object level. The monitoring process allows the meta controller to track the status of object-level processes and actions, as well as the consequences of those actions. However, current meta-reasoning, primarily as a resource manager, is insufficient for exhibiting intelligent and generalist behaviors, necessitating a meta-cognitive approach.
\subsection*{Graph Neural Networks}
A graph $ \mathcal{G} = (\mathcal{V}, \mathcal{E})$ is defined by a set of nodes $\mathcal{V}$ with node attributes $\mathtt{X}_v$ for $ v \in \mathcal{V}$ and a set of edges $\mathcal{E}$ between nodes. Edge from node $u \in \mathcal{V}$ to node $v \in \mathcal{V}$ is given by $ (u,v) \in \mathcal{E}$ with attributes $\mathtt{e}_{u,v}$.  One typical task is to predict the label or value of the entire graph.  
Graph neural networks (GNN), \cite{gilmer2017neural,hamilton2017inductive} and others, use node connectivity as well as node and edge features to learn a permutation-invariant embedding vector for each node $h_v , v \in \mathcal{V}$ and the entire graph $h_{\mathcal{G}}$. 
Typically, the representation is learned through message passing or neighborhood feature aggregation, in which node embedding vectors are updated in each layer based on message aggregation from neighbors:
\begin{equation}
h_v^{(k)} = \mathcal{F}' \left( h_v^{(k-1)}, \bigoplus_{u \in \mathcal{N}(v)} \mathcal{F}'' \left( h_v^{(k-1)}, h_u^{(k-1)}, \mathtt{e}_{u,v} \right) \right)
\end{equation}
where $h_v^{(k)}$ is the updated representation of node $v$ embedding in layer $k$, $\mathcal{N}(v)$ is the set of neighboring nodes of node $v$, $h_u^{(k-1)}$ is the representation of neighboring node $u$ in layer $k-1$, $\bigoplus$ is a differentiable permutation invariant aggregation function, $\mathcal{F}'$ and $\mathcal{F}''$ are differentiable functions usually neural networks. Node features are transformed prior to aggregation, and the methods used to transform and aggregate features differ between graph network variants. Motivated by the attention mechanism \cite{vaswani2017attention} in the transformer model, \cite{velivckovic2017graph} employs self-attention over node features and aggregating over neighborhood nodes in graph attention networks (GAT). We use the GATv2 operator \cite{brody2021attentive}, which improves static attention problems in graph attention layers.
\subsection*{Policy Gradient Methods}
We consider a standard reinforcement learning setting \cite{sutton2018reinforcement}
in which an agent interacts with the environment over a number of steps. A Markov decision process (MDP) is defined as a tuple $(\mathcal{S}, \mathcal{A}, \mathcal{P}, r, \gamma) $ in which $\mathcal{S}$ is a finite state space, $\mathcal{A}$ is a finite discrete action space, $\mathcal{P}: \mathcal{S} \times \mathcal{A} \times \mathcal{S} \rightarrow \mathbb{R} $ is the transition model, $ r : \mathcal{S} \times \mathcal{S} \rightarrow \mathbb{R} $ is the reward function and $\gamma \in (0, 1) $ is the discount factor. At each step t, the agent takes an action $a_t \in \mathcal{A}$ in state $s_t \in \mathcal{S}$ according to its policy $\pi$, which is a mapping of states to actions. As a result, the agent reaches the state $ s_{t+1} \in \mathcal{S} $, receives a scalar reward $r_t \in \mathbb{R}$, and accumulates returns;  $ R_t = \sum_{k=0}^{\inf}\gamma^k r_{t + k + 1}$ represents the accumulated return from step t. The agent's goal is to maximize the expected return for each state. 
The state-action value function $Q^{\pi}: \mathcal{S} \times \mathcal{A} \rightarrow \mathbb{R}$ is defined as $Q^{\pi}(s,a) = \mathbb{E} [R_t | s_t = s, a ]$, which is the expected return from selecting action  $a$ in state $s_t$ following policy $\pi$. 
The value of state $s$ under policy $\pi$ is defined as $V_{\pi}(s) = \mathbb{E}[R_t | s_t = s]$, which is the expected return from the state $s$.
Policy gradient methods, like REINFORCE \cite{williams1992simple}, directly learn a parameterized policy $\pi(s; \theta)$  and update its parameters $\theta$ using approximate gradient ascent of a performance measure, $ \nabla_\theta J(\theta) = \mathbb{E}_\pi [G_t \nabla_\theta \ln \pi_\theta(A_t \vert S_t)] $. 
To reduce the variance in the expected return estimate, a baseline is subtracted from it; typically, a learned value function is used as the baseline.
The resulting quantity $ A(s, a) = Q(s, a) - V(s) $ can be viewed as an estimate of the advantage of action $a$ in state $s$. 
This method is called advantage actor critic (A2C) \cite{sutton2018reinforcement,degris2012model}, and the agent acting according to the policy $\pi$ is the actor, while the value baseline is the critic.  
Several improved methods have been developed to address sample efficiency, stability, and training difficulty. 
Policies are learned by adapting Proximal policy optimization (PPO) \cite{schulman2017proximal}, an on-policy policy optimization that addresses concerns in actor-critic methods using clipped objectives. 
\section*{Metacognitive Learning Framework}
\label{sec:kix_model}
\begin{figure}[!ht]
    \centering
    \includegraphics[width=0.95\linewidth]{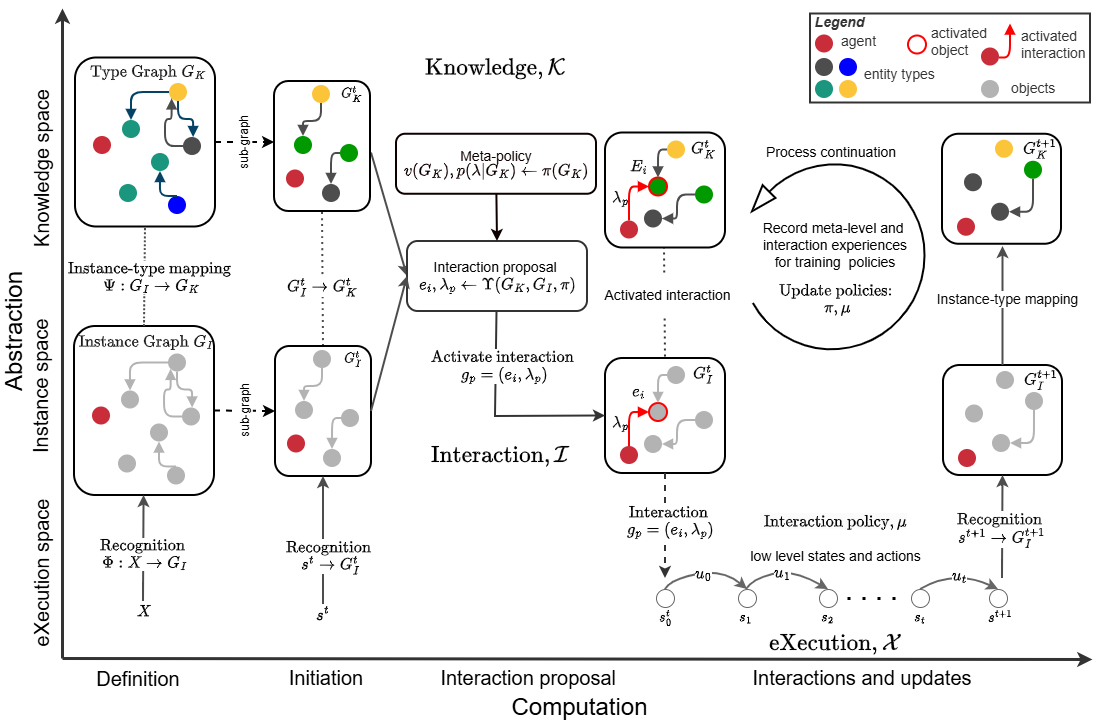}
    \vspace*{-1mm}
    \caption{
    The metacognitive learning framework comprises three levels: knowledge or meta ($\mathcal{K}$), interaction ($\mathcal{I}$), and execution ($\mathcal{X}$), based on abstraction and computation. The knowledge level contains type space knowledge graphs that represent higher-order abstractions, while the instance space comprises object instance graphs. The interaction level serves as an intermediary, making recommendations for interactions that the agent executes at the execution level. Computation proceeds through cycles of recognition, recommendation, execution, and interaction, to collect experiences for updates to both the meta-level and interaction policies.
    }
 \label{fig:kix_computation}
 \vspace*{-5mm}
\end{figure}

Our proposal for metacognitive learning centers on the interpretation of abstract knowledge to guide interactions with objects. A meta-policy recommends localized agent–object interactions, which the agent executes via interaction policies to construct a compositional task solution. Here, we outline the main ideas of the metacognitive learning framework, Knowledge-Interaction-eXecution (KIX) (see Figure~\ref{fig:kix_computation}).
\subsection*{Knowledge}
The environment $\Xi$ is partially observable and consists of objects $e = \{e_1, e_2,...,e_n\}$ and an agent $A$.
It is assumed that the agent has a recognition system that can access its partial view $X$ and associate objects with their corresponding types, such that the type of an object $e_i \in e$ is in entity types $E = \{E_1, E_2, ... , E_m\}$.
The agent can perform low-level actions $u=\{u_1, u_2, ... \}$ within the environment.
When making decisions, the agent organizes its knowledge into instance, and type graphs.
An instance graph is an ordered pair $G_I = (e, \Lambda)$ of observed objects $e$ and their relations $\Lambda = \{\Lambda_1,\Lambda_2,...,\Lambda_{\kappa} \}$.
A type graph is an ordered pair $G_K = (E, \Lambda)$ of entity types $E$ and their abstract relations $\Lambda$. Relations, whether between objects or entity types, retain the same meaning, such as adjacency indicating that two objects are close to one another.
The function $\Phi: X \to G_I$ maps the agent's partially observed view to the instance graph by capturing observed objects and relations. 
The function $\Psi: G_I \to G_K$ maps instance graphs to type graphs, a direct translation since the agent's recognition system contains types corresponding to the objects.
If the agent encounters a novel object, it must create a new type in its type space or assign the object to a related type. In the framework, $\Psi$ is learnable and represents a belief, mapping objects to types. However, this article treats the mapping as fixed, from object nodes in the instance graph $G_I$ to type nodes in the type graph $G_K$.
As the agent interacts with objects, the environment changes, and its recognition system captures a new partially observed view, resulting in new instance and type graphs. 

The agent's knowledge level is also referred to as the meta-level. At the meta-level, the agent uses a meta-policy that accepts a type graph and returns interactions or meta-level actions for object interaction. Interactions are represented as $\lambda = \{\lambda_1,\lambda_2,...,\lambda_p \}$.
For each interaction type, the agent uses an interaction policy to execute low-level actions $u$ within the environment. Therefore, the agent's interaction level 
$\mathcal{I} = (A, G_I, \lambda)$ comprises its instance graph and available interactions, while the knowledge or meta level $\mathcal{K} = (A, G_K, \lambda)$ comprises its type graph and meta level actions or interactions.
\subsection*{Interaction}
After the agent constructs the type and instance graphs, the meta-level recommender $\Upsilon$ recommends a candidate object $e_i$ and interaction type $\lambda_p$ for an interaction based on the agent's type and instance graphs and meta-level policy $\pi$.
The recommender iterates through all visible objects in the agent's instance graph $G_I$. For each object, it adds an activation relation between the agent and the object in both the instance graph $G_I$ and the type graph $G_K$, where `activating' refers to observing or focusing on an object for potential interaction.
For $G_K$, the agent's meta-policy returns the state value and the interaction probability. The state value represents the value of activating that specific object. Then, an object $e_i$ for interaction is proposed by sampling based on the values of activation of the visible objects, and interaction $\lambda_p$ is sampled from the interaction probability of the proposed object. Therefore, at each step, the meta-level of the agent activates a candidate object $e_i$ for interaction $\lambda_p$, setting the agent's interaction goal $g^i_p = (e_i, \lambda_p)$.
Acting on the recommended interaction goal, the agent begins interacting with the environment. After each round of interaction, the interaction status is passed to the meta-level, which then proposes the next candidate object and interaction. This process continues until the task is completed or the environment is terminated.
\subsection*{eXecution}
Each agent has a meta-level policy, whereas interaction policies are a collection of policies corresponding to different interactions. 
By following the interaction goals $g^i_p$ as directed by the meta-level, the agent interacts with objects. During an interaction $\lambda_p$, the agent executes a series of low-level actions $u$ on execution level states $s$ based on the relevant interaction policy $\mu_p$.
For example, if there is a door, the agent can activate it and interact with it, like `open' the door, even if this requires multiple low-level actions such as moving and toggling.
As the agent interacts with objects, it collects trajectories of experiences at both the meta and interaction levels. The meta-level trajectory is used to learn the meta-level policy, while the low-level interaction experiences are used to learn the collection of interaction policies. 
The learning approach for the meta-policy and interaction policies is detailed below.
\subsection*{Meta-policy Learning}
Meta-level representations of type graphs with activation relations are learned using a message-passing graph neural network. Nodes have abstract level attributes such as type encoding, while edges have edge-type encoding for different types of relations. We use GAT convolutional layers \cite{brody2021attentive,fey2019fast} to learn graph representations given as,
\begin{equation}
h_v' = \alpha_{v,v} \theta  h_v + \sum_{u \in \mathcal{N}(v)} \alpha_{u,v} \theta  h_u
\end{equation}
where $h_v'$ is the updated embedded representation of node $v$, $\mathcal{N}(v)$ is the set of neighbors of node $v$, and $h_u$ is the representation of the neighboring node. $\theta$ are learnable weights shared by source and target nodes in the neighborhood, and the attention coefficients $\alpha_{u,v}$ are computed using weighted node features.

The hidden representation after the final graph attention layer is pooled for graph rollout, which is then passed to the fully connected actor and critic layers of the model.
The meta policy $\pi_{\theta}(.| G_K )$, parameterized by $\theta$, returns a state value and an interaction or meta-action, $\lambda$, distribution.
The recommender selects a target object and an interaction $\lambda_p$, and the agent works through low-level interactions based on guidance.
At the end of the interaction, the status of the interaction is returned to the meta level, and the agent collects the trajectory of meta-level experiences, $\mathcal{D}_{\tau_m}$.
An estimate of advantages is computed as $ A_m = r_m + \gamma_m V_{\pi_{\theta}}(G_K') - V_{\pi_{\theta}} (G_K) $.

Meta level PPO optimizes a surrogate loss:
\begin{equation}
\mathcal{L}_m^{\text{actor}}(\theta) = - \mathbb{E}_{\mathcal{D}_{\tau_m}} \left[ \min\left(r_{\theta} A_m, \text{clip}(r_{\theta}, 1-\epsilon, 1+\epsilon) A_m \right) \right] 
\end{equation}
with $ r_{\theta} = \frac{\pi_{\theta}( . | G_K )}{\pi_{\theta}^r ( . | G_K )} $ being the ratio of probabilities of actions at meta level roll-out. The critic loss is the mean squared error between the estimated values and the returns:
\begin{equation}
\mathcal{L}_m^{\text{critic}}(\theta) = \mathbb{E}_{\mathcal{D}_{\tau_m}} \left[ \frac{1}{2} (V_{\pi_{\theta}}(G_K) - R_m)^2 \right] 
\end{equation}
Considering the mean entropy $ \mathcal{H}_{\theta} $, the overall minimization objective at the meta-level becomes:
\begin{equation}
\mathcal{L}_m(\theta) = \mathcal{L}_m^{\text{actor}}(\theta) - \zeta_m^h \mathcal{H}_{\theta} + \zeta_m^c \mathcal{L}_m^{\text{critic}}(\theta) 
\end{equation}
where $\zeta_m^h$, $\zeta_m^c$ are the entropy and value coefficients, respectively.
The model parameters $\theta$ are optimized as $ \theta \leftarrow \theta + \alpha_m \nabla_{\theta} L_m(\theta) $, and the updated meta-policy is used for the next episode of learning.
\subsection*{Interaction Policy Learning}
Interaction policies are a repository of policies for different interactions which can be recommended by the meta level, $\mu_{\phi} = \{ \mu_{\phi_1}, \mu_{\phi_2}, ... , \mu_{\phi_p} \} $ and each policy is parameterized separately.
When an interaction goal with an object $ e_i $ and interaction $\lambda_p$ is recommended, $g^i_p = (e_i, \lambda_p)$, the agent attempts to achieve it by executing a series of low-level actions from $u$. 
The agent observes a low-level state $s$ from a convolutional neural network embedding, samples an action $ u \sim \mu_{\phi_p} ( . | s ) $ using the respective interaction policy, acts in the environment, and collects the trajectory of experiences $\mathcal{D}_{\tau_p}$.
The experiences collected are used to learn interaction policies.
An estimate of advantages is computed as $ A_p = r_p + \gamma_p V_{\mu_{\phi_p}}(s') - V_{\mu_{\phi_p}} (s) $.

Each interaction level PPO optimizes a surrogate loss:
\begin{equation}
\mathcal{L}_p^{\text{actor}}(\phi_p) = -\mathbb{E}_{\mathcal{D}_{\tau_p}}\left[ \min\left(r_{\phi_p} A_p, \text{clip}(r_{\phi_p},1-\epsilon,1+\epsilon) A_p \right) \right]
\end{equation}
with $ r_{\phi_p} = \frac{\mu_{\phi_p} ( . | s )}{ \mu_{\phi_p}^r ( . | s )} $ is the ratio of probabilities of actions of policy and roll-out policy.  The critic loss is the mean squared error between the estimated values and the returns:
\begin{equation} 
\mathcal{L}_p^{\text{critic}}(\phi_p) = \mathbb{E}_{\mathcal{D}_{\tau_p}} \left[ \frac{1}{2} (V_{\mu_{\phi_p}}(s) - R_p)^2 \right] 
\end{equation}
Considering the mean entropy $ \mathcal{H}_{\phi_p} $,  the overall minimization objective becomes:
\begin{equation} 
\mathcal{L}_p(\phi_p) = \mathcal{L}_p^{\text{actor}}(\phi_p) - \zeta_p^h \mathcal{H}_{\phi_p} + \zeta_p^c \mathcal{L}_p^{\text{critic}}(\phi_p) 
\end{equation}
for each interaction $p$, where $\zeta_p^h$, $\zeta_p^c$ are the entropy and value coefficients, respectively.
For each interaction policy, the model parameters $\phi_p$ are optimized as $ \phi_p \leftarrow \phi_p + \alpha \nabla_{\phi_p} L_p(\phi_p) $, and the updated interaction policies are used for the next episode of learning.

In summary, when viewing tasks as meta-policy guided interactions with objects, problems are naturally decomposed into interaction goals. Unlike the models that are tightly coupled with low-level representations through end-to-end policies, in the proposed framework, agent actions are aided by the interpretation of meta-level knowledge to guide object interactions, providing a metacognitive mechanism for knowledge learning, abstract reasoning and generalist behaviors.
\section*{Results and Discussion}
\label{sec:kix_results}
\subsection*{Experiments}
\begin{figure}[ht]
    \vspace*{-3mm}
    \centering
    \begin{subfigure}[c]{0.495\linewidth}
        \centering
        \includegraphics[width=0.405\textwidth]{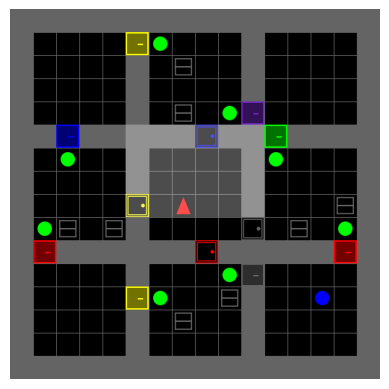}
        \subcaption*{\small{ObstructedMaze-Full}}
    \end{subfigure}
    \begin{subfigure}[c]{0.495\linewidth}
        \centering
        \includegraphics[width=0.405\textwidth]{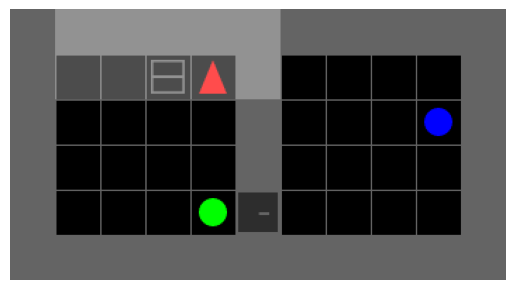}
        \subcaption*{\small{ObstructedMaze-1Dlhb}}
    \end{subfigure}
    \vspace*{-1mm}
    \caption{
    \emph{Environments}: ObstructedMaze-Full is used to experimentally evaluate the agent's knowledge learning and flexible task solving capabilities, while ObstructedMaze-1Dlhb is used to evaluate the transferability of interaction policies.
    }
    \label{fig:kix_environment}
    \vspace*{-5mm}
\end{figure}

We aim to examine whether agents can learn to associate abstract relational knowledge with object and interaction types, leverage this knowledge for interaction learning and guide interactions to exhibit flexible behavior when solving complex tasks.
To assess the proposed agents, we need a challenging environment having ingredients for higher-order reasoning to solve tasks. MiniGrid world navigation environments \cite{chevalier2018minimalistic} are suitable for this purpose; they are procedurally generated environments where the agent interacts with objects such as doors, keys, balls, and boxes. The obstructed maze environments, shown in Figure~\ref{fig:kix_environment}, are particularly challenging due to partial observability, stochasticity, and sparse rewards, with only a few claims for solving it. In the MiniGrid-ObstructedMaze-Full environment, the agent is initially placed in the middle room, with a 7x7 view. At the low level, the agent can move forward, left, right, pick, drop, and toggle, and its goal is to pick up the blue ball, which is randomly placed in one of the corner rooms.  Some of the doors are locked and require a key of the same color to open. Keys are hidden inside boxes, but the agent cannot see inside the boxes without toggling. 
The default task is referred to as Task 0. Furthermore, we identify the following new task scenarios, which were presented to the agent during testing but not during training:
\begin{itemize}
    \item No context in training (Task 1) -- a relation that the agent did not come across during training.  It is simulated by placing the goal inside an empty box. During training, the blue ball was never in a box, although the agent encountered boxes and balls during training, but the boxes contained keys.
    \item Different context in testing (Task 2) -- using a relation that the agent encountered during training but in a new context during testing. It is simulated by locking doors in the central room and placing keys directly in the room. During training, the doors in the central room were never locked, nor was there direct access to the keys.
    \item Dynamic changes in the environment (Task 3) -- when some event occurs, the environment changes. During training, the goal location in an episode remained unchanged. To simulate this, as soon as the agent enters the room with the goal, relocate the goal to one of the neighboring rooms. It is done only once, not repeatedly.
\end{itemize}
\subsection*{Knowledge Learning to Interact}
We devised two variants: KIX-A and KIX-R. Both variants learn a meta-policy and a collection of interaction policies. 
While the KIX-A variant combines both reaching objects and interacting with them, the KIX-R variant learns an additional interaction policy referred to as the `reachability' policy, which is used to reach objects prior to interacting with them. 
All policies, including meta and interactions, are implemented as actor-critic PPO policies.

\vspace*{-3mm}
\begin{algorithm}
    \DontPrintSemicolon
    \SetAlgoLined
    \SetNoFillComment
    \LinesNotNumbered
    \caption{KIX procedure}
    \label{alg:kix}
    initialize model weights $\theta, \phi$ , parameters, $T$ \;
    \For{ $t = 1$ \KwTo $T$}{
    \Comment{collect meta experiences}
    \For{$k = 1$ \KwTo $K$}{
        observe a partial view $X$ \;
        make instance graph $G_I \leftarrow \Phi(X)$ \;
        map $G_I$ to $G_K$, $G_K \leftarrow \Psi(G_I)$ \;
        recommend $(e_i,\lambda_p) \leftarrow \Upsilon(G_K,G_I,\pi)$ \;
        interact with object $e_i$ following $\mu_{\lambda_p}$ \;
        collect interaction exps $\mathcal{D}_{\tau_p}$ and compute $A_p$\; 
    }
    collect meta exp $\mathcal{D}_{\tau_m}$ and compute $A_m$ \;
    \Comment{Update policies}
    update $\pi$ by optimizing $\mathcal{L}_m(\theta) $ \;
    \For{ $\lambda_p$}
    { update $\mu_{\lambda_p}$ by optimizing $\mathcal{L}_p(\phi_p) $ \; }
    }
\end{algorithm}
\vspace*{-3mm}
%

Algorithm~\ref{alg:kix} provides an overview of the procedure. The agent's partially observed view $X$ contains objects with identifiers and attributes such as color and type.
The function $\Phi$ takes $X$ and creates an instance space graph by using visible objects as nodes, including the agent, and determining relationships between nodes such as `visible', `adjacent', and `carrying'. The `adjacent' relation is derived using distance between the two objects.  When the agent chooses an object to interact with, an ‘activated' relation is added between the agent and the object node. The function $\Psi$ then constructs a type space graph from the instance space graph, encoding node types as attribute vectors.  Edges in a type space graph have the same meaning as those in an instance space graph and are encoded as one-hot encoded edge attribute vectors. 

Agents learn how to perform low-level actions in the execution space using interaction policies. A convolutional network with three Conv2D layers with 16, 32, and 64 output channels, a max pooling layer added after the first convolution layer, and exponential linear units as activation functions is used to provide the input state embeddings for the interaction actor-critic. An activation indicator for the proposed interaction object is added as a new input channel, resulting in a four-channel input to the filters. Actor-critic networks are fully connected with a single hidden layer. 
Agents learn a meta-policy in the type space. Two graph attention layers with 4 heads and 16 dimensional embedding provide 64 dimensional input embeddings for the meta policy. During message passing, these layers share weights between source and target. The final graph attention layer is then used for graph readout, applying a global add pool. The meta-level actor and critic are both fully connected networks, with the critic learning a single value and the actor learning the likelihood of taking the meta-level actions `pickup', `drop’, `reveal', `open', or `open with key', each of which corresponds to an interaction policy. 
\begin{table}[h]    
    \vspace*{-3mm}
    \centering
    \begin{tabular}[c]{@{}ll|ll|ll@{}}    
        \toprule
        \textbf{Parameter} & \textbf{Value} & \textbf{Parameter} & \textbf{Value} & \textbf{Parameter} & \textbf{Value} \\
        \midrule
        meta exp size        & 8      &  learning rate       & 1e-05  & gamma       & 0.99 \\
        exp size             & 64     &  entropy coeff       & 1e-04  & gae lambda  & 0.95 \\
        meta minibatch size  & 64     &  value coeff         & 1.0    & ppo clip    & 0.2  \\
        minibatch size       & 64     &  meta entropy coeff  & 1e-04  & ppo steps   & 4    \\
        meta learning rate   & 1e-05  &  meta value coeff    & 1.0    &             &      \\
        \bottomrule
    \end{tabular}
    \vspace*{2mm}
    \caption{Training hyper-parameters}
    \label{table:kix_net}
    \vspace*{-5mm}
\end{table}
The environment rewards the agents with $r=1-(c/c_T)$ where $c$ is the number of steps taken and $c_T$ is the max possible steps.  For interaction policies, the agent receives a reward of 1 when it has a successful interaction. At the meta-level, it is rewarded with 0.1 for a successful proposed interaction and $1+r$ for achieving the environment goal. 
The base agent, BASE, is an end to end policy learner with a similar actor-critic architecture and receives environment rewards. 
The training hyper-parameters are given in Table \ref{table:kix_net}. Learning rates were determined using grid search with values 1e-3, 1e-4, 1e-5, and 1e-6. 
All the agents are trained for $\sim 6.2 \times 10^8$ steps, using 32 parallel workers in a high performance computing cluster.
%
\begin{figure}[!htp]
        \centering
        \begin{subfigure}[c]{0.325\linewidth}
            \centering
            \includegraphics[width=0.99\textwidth]{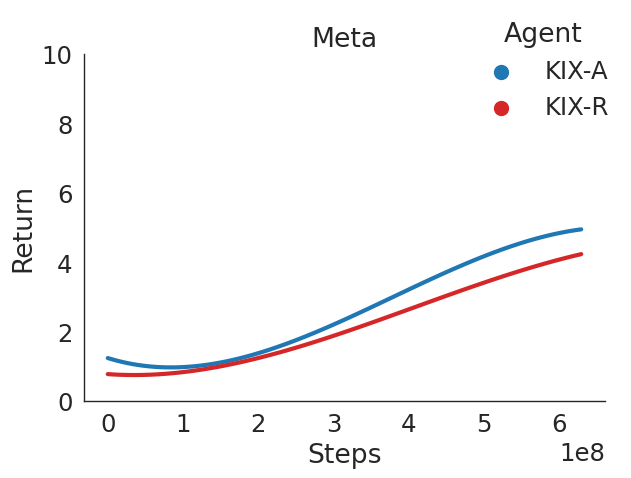}
        \end{subfigure}
        \hfill
        \begin{subfigure}[c]{0.325\linewidth}
            \centering
            \includegraphics[width=0.99\textwidth]{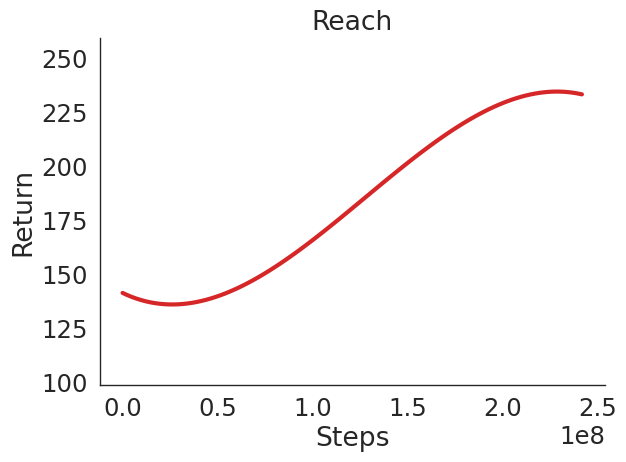}
        \end{subfigure}
        \hfill
        \begin{subfigure}[c]{0.325\linewidth}
            \centering
            \includegraphics[width=0.99\textwidth]{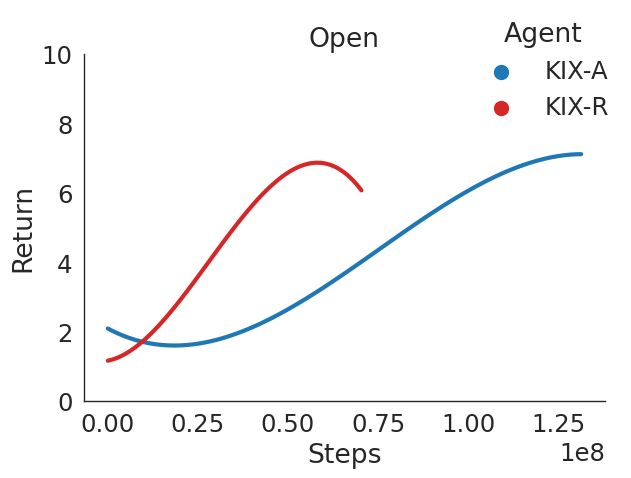}
        \end{subfigure}
        \vfill
        \begin{subfigure}[c]{0.325\linewidth}
            \centering
            \includegraphics[width=0.99\textwidth]{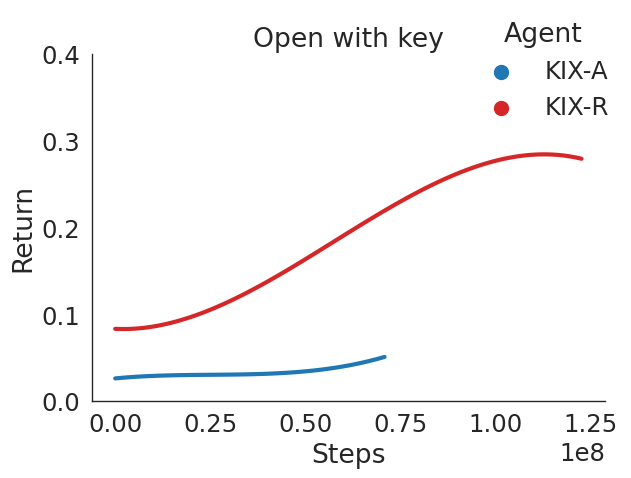}
        \end{subfigure}
        \hfill        
        \begin{subfigure}[c]{0.325\linewidth}
            \centering
            \includegraphics[width=0.99\textwidth]{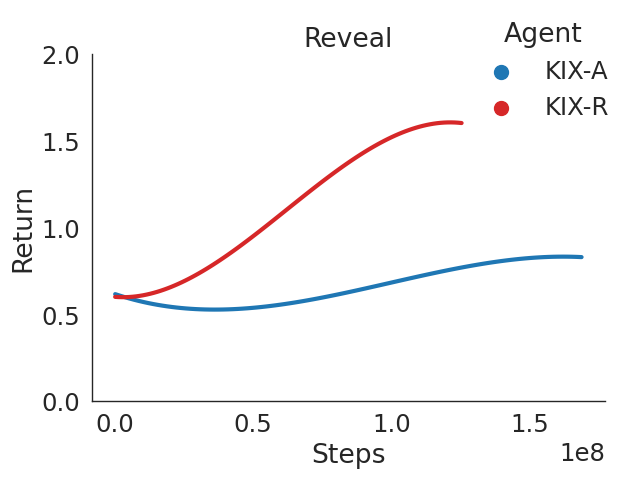}
        \end{subfigure}
        \hfill
        \begin{subfigure}[c]{0.325\linewidth}
            \centering
            \includegraphics[width=0.99\textwidth]{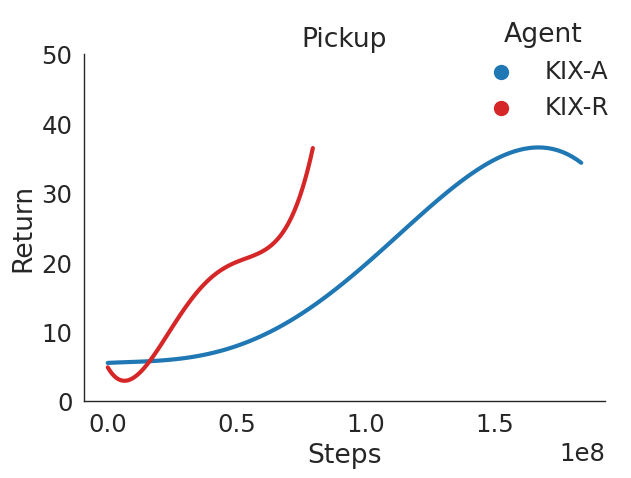}
        \end{subfigure}
    \caption{
    \emph{Agents learning to interact }: The learning progress of both the agents KIX-A (blue) and KIX-R (red) is shown as returns over steps, demonstrating the learning of meta-level as well as interaction policies such as `reach' (only applicable for KIX-R), `open', `open with key', `reveal', `pickup', `drop' (not shown) in the ObstructedMaze-Full environment.
    }
    \label{fig:kix_training}
    \vspace*{-3mm}
\end{figure}
%
As shown in Figure~\ref{fig:kix_training}, as training progresses, agents learn to interact with objects and associate higher-level structured knowledge with these interactions using object types and type and instance graphs, as indicated by higher meta-level returns. This, in turn, leads to better interaction experiences and further improvements in interaction policies. 
Furthermore, the KIX-R variant first focuses on learning how to reach different objects, but once it learns to reach, it begins to learn interactions in fewer steps and with much higher returns compared to the KIX-A variant.
It suggests that the agents learn higher-order relations and apply these learned relations in interactions with objects to solve problems.
\subsection*{Meta-level Guided Interactions Aid Agents}
Our approach is motivated by the idea that meta-level guided interactions with objects help agents in generalizing while performing tasks, rather than acting directly at the execution level. Since the agent's environmental return is determined by the number of steps taken to complete a task, completing tasks in fewer steps yields a higher return. Therefore, the distribution of environmental returns is a reliable indicator of the value of meta-level interaction recommendations for agents. We evaluate the distribution of environmental returns for each agent across four tasks. 
\begin{wrapfigure}{r}{0.5\textwidth}
    \vspace*{-3mm}
    \centering
    \includegraphics[width=0.99\linewidth]{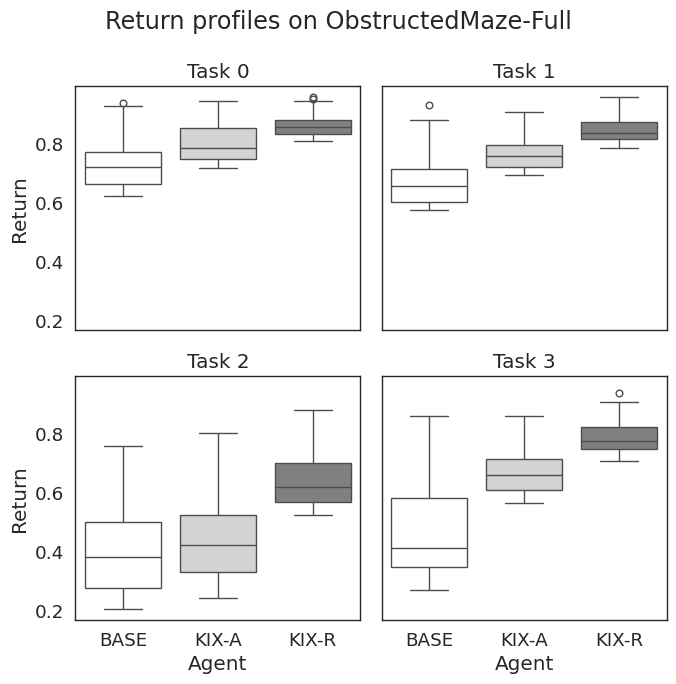}
    \caption{
    Evaluation return profiles of agents performing tasks show that the agents benefit from meta-level guided interactions.
    }
    \label{fig:kix_metrics1a}
    \vspace*{-3mm}
\end{wrapfigure}
For each evaluation task, trained agents are run for 12K episodes by parallel workers, and their environmental rewards are tracked. 
To understand the effects of meta-guided interactions and ensure a fair comparison between agents, regardless of success rate differences, we examine the top $k$ returns with $k=100$. Figure~\ref{fig:kix_metrics1a} shows the return profiles of agents. It is evident from the distribution of returns that the proposed agents achieve higher returns on all tasks compared to the base agent, including dynamic tasks like Task 3. 
This demonstrates that meta-guided interactions not only assist agents in completing tasks in fewer steps, but also enable them to perform a wider range of tasks in both static and dynamic scenarios. 
Furthermore, higher returns of the second variant show that agents benefit from treating reachability as a separate interaction. 
In the absence of separate reachability, reaching is embedded into interactions. Separate reachability allows shared learning followed by interactions, which is particularly beneficial in dynamically shifting scenarios.
\subsection*{Transferability of Policies}
%
\begin{figure}[!ht]
    \vspace*{-3mm}
    \centering
    \begin{subfigure}[b]{0.495\linewidth}
        \centering
        \includegraphics[width=0.7\textwidth]{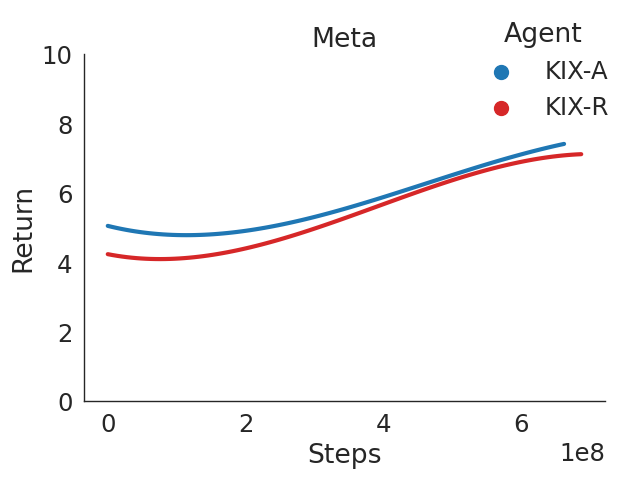}
        \vspace*{-2mm}
        \caption{Meta training}
    \end{subfigure}
    \hfill
    \begin{subfigure}[b]{0.495\linewidth}
        \centering
        \includegraphics[width=0.7\textwidth]{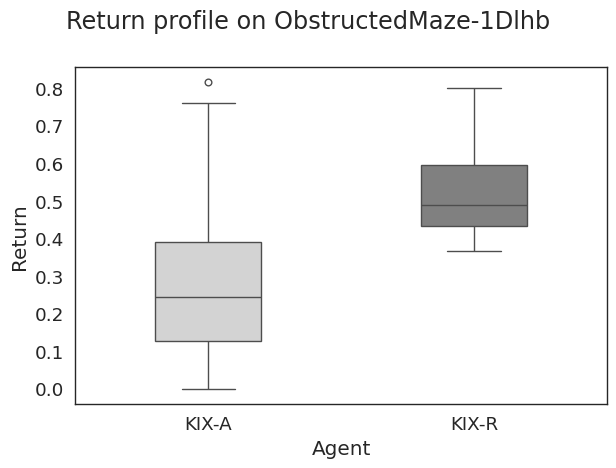}
        \vspace*{-2mm}
        \caption{Evaluation of task returns}
    \end{subfigure}
    %
    \caption{
    Evaluation return profiles show that interaction policies can be transferred from one environment to another.
    }
    \label{fig:kix_metrics1b}
    \vspace*{-3mm}
\end{figure}
The ObstructedMaze-1Dlhb environment is used to assess the transferability of interaction policies learned in one environment to another. In this environment, the agent's goal is to pick up a blue ball hidden in a two-room maze. The rooms are separated by a locked door that is obstructed by a ball. The keys are hidden inside boxes. Agents employ previously acquired interaction policies from the base task in the ObstructedMaze-Full environment while only learning the meta-policies. In evaluation, agents use the learned meta-policy, but the interaction policies are from the ObstructedMaze-Full environment. 
Figure~\ref{fig:kix_metrics1b} shows evaluation return profiles for the base task, similar to the main environment, for both agent variants. 
It is evident that agents have gained the ability to exploit interaction policies from the previous environment, and the second variant, which employs a separate reachability policy, appears to have benefited more.
\section*{Conclusion}
\label{sec:conclusion}
Artificial intelligence systems must develop knowledge-centric metacognitive abilities in the pursuit of abstract reasoning, knowledge learning, adaptive behavior and autonomy. The proposed metacognitive approach promotes these capabilities through three core contributions.
\emph{First}, knowledge learning and adaptation necessitate abstraction and composition while preserving relational knowledge structures. In our model, type graphs are used to represent and ground natural abstractions while preserving relational knowledge among objects.
\emph{Second}, 
rather than acting directly in the environment, interactions with objects are essential in task execution. To accomplish this, we introduced an intermediary interaction level. At the meta level, the agent interprets abstract relational knowledge, learns a meta-policy over type graphs, and generates recommendations regarding which objects to interact with and how. In this sense, the proposed model functions as an interpretation-based interaction goal generating machine.
\emph{Third}, 
meta level guided interactions executed through interaction policies provide the loose coupling necessary for generalization by acting as intermediaries. The metacognitive approach facilitates knowledge learning of policies, knowledge guided interactions aid generalization and transferable interaction policies.
\emph{Lastly}, 
the proposed framework is modular and adaptable, enabling the integration of knowledge structures and control variables at both the meta and interaction levels. It is particularly promising for physical-world domains where abstract relational knowledge and compositional interactions can be exploited, such as robotics, autonomous, and cognitive systems.
%
\balance
\bibliographystyle{named}
\bibliography{kix}

@book{bishop2006pattern,
  title     = {Pattern recognition and machine learning},
  author    = {Bishop, Christopher M and Nasrabadi, Nasser M},
  volume    = {4},
  year      = {2006},
  publisher = {Springer}
}

@book{Goodfellow2016,
  title     = {Deep Learning},
  author    = {Ian Goodfellow and Yoshua Bengio and Aaron Courville},
  publisher = {MIT Press},
  note      = {\url{http://www.deeplearningbook.org}},
  year      = {2016}
}

@inproceedings{veerapaneni2020entity,
  title={Entity abstraction in visual model-based reinforcement learning},
  author={Veerapaneni, Rishi and Co-Reyes, John D and Chang, Michael and Janner, Michael and Finn, Chelsea and Wu, Jiajun and Tenenbaum, Joshua and Levine, Sergey},
  booktitle={Conference on Robot Learning},
  pages={1439--1456},
  year={2020},
  organization={PMLR}
}

@article{campero2020learning,
  title={Learning with amigo: Adversarially motivated intrinsic goals},
  author={Campero, Andres and Raileanu, Roberta and K{\"u}ttler, Heinrich and Tenenbaum, Joshua B and Rockt{\"a}schel, Tim and Grefenstette, Edward},
  journal={arXiv preprint arXiv:2006.12122},
  year={2020}
}

@article{vijay2019generalization,
  title={Generalization to novel objects using prior relational knowledge},
  author={Vijay, Varun Kumar and Ganesh, Abhinav and Tang, Hanlin and Bansal, Arjun},
  journal={arXiv preprint arXiv:1906.11315},
  year={2019}
}

@article{dayan1992feudal,
  title   = {Feudal reinforcement learning},
  author  = {Dayan, Peter and Hinton, Geoffrey E},
  journal = {Advances in neural information processing systems},
  volume  = {5},
  year    = {1992}
}

@article{sutton1999between,
  title     = {Between MDPs and semi-MDPs: A framework for temporal abstraction in reinforcement learning},
  author    = {Sutton, Richard S and Precup, Doina and Singh, Satinder},
  journal   = {Artificial intelligence},
  volume    = {112},
  number    = {1-2},
  pages     = {181--211},
  year      = {1999},
  publisher = {Elsevier}
}

@article{pateria2021hierarchical,
  title     = {Hierarchical reinforcement learning: A comprehensive survey},
  author    = {Pateria, Shubham and Subagdja, Budhitama and Tan, Ah-hwee and Quek, Chai},
  journal   = {ACM Computing Surveys (CSUR)},
  volume    = {54},
  number    = {5},
  pages     = {1--35},
  year      = {2021},
  publisher = {ACM New York, NY, USA}
}

@inproceedings{bagaria2021skill,
  title        = {Skill discovery for exploration and planning using deep skill graphs},
  author       = {Bagaria, Akhil and Senthil, Jason K and Konidaris, George},
  booktitle    = {International Conference on Machine Learning},
  pages        = {521--531},
  year         = {2021},
  organization = {PMLR}
}

@article{barto2003recent,
  title     = {Recent advances in hierarchical reinforcement learning},
  author    = {Barto, Andrew G and Mahadevan, Sridhar},
  journal   = {Discrete event dynamic systems},
  volume    = {13},
  pages     = {341--379},
  year      = {2003},
  publisher = {Springer}
}

@article{dietterich2000hierarchical,
  title   = {Hierarchical reinforcement learning with the MAXQ value function decomposition},
  author  = {Dietterich, Thomas G},
  journal = {Journal of artificial intelligence research},
  volume  = {13},
  pages   = {227--303},
  year    = {2000}
}

@article{hitzler2022neuro,
  title={Neuro-symbolic artificial intelligence: The state of the art},
  author={Hitzler, Pascal and Sarker, Md Kamruzzaman},
  year={2022},
  publisher={IOS press}
}

@article{zhang2024neuro,
  title={Neuro-Symbolic AI: Explainability, Challenges, and Future Trends},
  author={Zhang, Xin and Sheng, Victor S},
  journal={arXiv preprint arXiv:2411.04383},
  year={2024}
}

@article{ackermanthompson2017meta,
  title     = {Meta-reasoning: Monitoring and control of thinking and reasoning},
  author    = {Ackerman, Rakefet and Thompson, Valerie A},
  journal   = {Trends in cognitive sciences},
  volume    = {21},
  number    = {8},
  pages     = {607--617},
  year      = {2017},
  publisher = {Elsevier}
}

@article{ten2022curiosity,
  title={Curiosity-Driven Exploration},
  author={Ten, Alexandr and Oudeyer, Pierre-Yves and Moulin-Frier, Cl{\'e}ment},
  journal={The Drive for Knowledge: The Science of Human Information Seeking},
  pages={53},
  year={2022},
  publisher={Cambridge University Press}
}

@article{oudeyer2007intrinsic,
  title={Intrinsic motivation systems for autonomous mental development},
  author={Oudeyer, Pierre-Yves and Kaplan, Frdric and Hafner, Verena V},
  journal={IEEE transactions on evolutionary computation},
  volume={11},
  number={2},
  pages={265--286},
  year={2007},
  publisher={IEEE}
}

@inproceedings{klyubin2005empowerment,
  title={Empowerment: A universal agent-centric measure of control},
  author={Klyubin, Alexander S and Polani, Daniel and Nehaniv, Chrystopher L},
  booktitle={2005 ieee congress on evolutionary computation},
  volume={1},
  pages={128--135},
  year={2005},
  organization={IEEE}
}

@inproceedings{portelas2020teacher,
  title={Teacher algorithms for curriculum learning of deep rl in continuously parameterized environments},
  author={Portelas, R{\'e}my and Colas, C{\'e}dric and Hofmann, Katja and Oudeyer, Pierre-Yves},
  booktitle={Conference on Robot Learning},
  pages={835--853},
  year={2020},
  organization={PMLR}
}

@article{hutsebaut2022hierarchical,
  title={Hierarchical reinforcement learning: A survey and open research challenges},
  author={Hutsebaut-Buysse, Matthias and Mets, Kevin and Latr{\'e}, Steven},
  journal={Machine Learning and Knowledge Extraction},
  volume={4},
  number={1},
  pages={172--221},
  year={2022},
  publisher={MDPI}
}

@article{chevalier2018minimalistic,
  title={Minimalistic gridworld environment for openai gym},
  author={Chevalier-Boisvert, Maxime and Willems, Lucas and Pal, Suman},
  journal={GitHub repository},
  year={2018}
}

@book{sutton2018reinforcement,
  title={Reinforcement learning: An introduction},
  author={Sutton, Richard S and Barto, Andrew G},
  year={2018},
  publisher={MIT press}
}

@inproceedings{degris2012model,
  title={Model-free reinforcement learning with continuous action in practice},
  author={Degris, Thomas and Pilarski, Patrick M and Sutton, Richard S},
  booktitle={2012 American Control Conference (ACC)},
  pages={2177--2182},
  year={2012},
  organization={IEEE}
}

@article{schulman2017proximal,
  title={Proximal policy optimization algorithms},
  author={Schulman, John and Wolski, Filip and Dhariwal, Prafulla and Radford, Alec and Klimov, Oleg},
  journal={arXiv preprint arXiv:1707.06347},
  year={2017}
}

@article{williams1992simple,
  title={Simple statistical gradient-following algorithms for connectionist reinforcement learning},
  author={Williams, Ronald J},
  journal={Machine learning},
  volume={8},
  pages={229--256},
  year={1992},
  publisher={Springer}
}

@inproceedings{gilmer2017neural,
  title={Neural message passing for quantum chemistry},
  author={Gilmer, Justin and Schoenholz, Samuel S and Riley, Patrick F and Vinyals, Oriol and Dahl, George E},
  booktitle={International conference on machine learning},
  pages={1263--1272},
  year={2017},
  organization={PMLR}
}

@article{hamilton2017inductive,
  title={Inductive representation learning on large graphs},
  author={Hamilton, Will and Ying, Zhitao and Leskovec, Jure},
  journal={Advances in neural information processing systems},
  volume={30},
  year={2017}
}

@article{velivckovic2017graph,
  title={Graph attention networks},
  author={Veli{\v{c}}kovi{\'c}, Petar and Cucurull, Guillem and Casanova, Arantxa and Romero, Adriana and Lio, Pietro and Bengio, Yoshua},
  journal={arXiv preprint arXiv:1710.10903},
  year={2017}
}

@article{brody2021attentive,
  title={How attentive are graph attention networks?},
  author={Brody, Shaked and Alon, Uri and Yahav, Eran},
  journal={arXiv preprint arXiv:2105.14491},
  year={2021}
}

@article{vaswani2017attention,
  title={Attention is all you need},
  author={Vaswani, Ashish and Shazeer, Noam and Parmar, Niki and Uszkoreit, Jakob and Jones, Llion and Gomez, Aidan N and Kaiser, {\L}ukasz and Polosukhin, Illia},
  journal={Advances in neural information processing systems},
  volume={30},
  year={2017}
}

@article{fey2019fast,
  title={Fast graph representation learning with PyTorch Geometric},
  author={Fey, Matthias and Lenssen, Jan Eric},
  journal={arXiv preprint arXiv:1903.02428},
  year={2019}
}

@incollection{nelson1990metamemory,
  title={Metamemory: A theoretical framework and new findings},
  author={Nelson, Thomas O},
  booktitle={Psychology of learning and motivation},
  volume={26},
  pages={125--173},
  year={1990},
  publisher={Elsevier}
}

@article{liu2022goal,
  title={Goal-conditioned reinforcement learning: Problems and solutions},
  author={Liu, Minghuan and Zhu, Menghui and Zhang, Weinan},
  journal={arXiv preprint arXiv:2201.08299},
  year={2022}
}

@article{pong2018temporal,
  title={Temporal difference models: Model-free deep rl for model-based control},
  author={Pong, Vitchyr and Gu, Shixiang and Dalal, Murtaza and Levine, Sergey},
  journal={arXiv preprint arXiv:1802.09081},
  year={2018}
}
\end{document}